# Why Do We Need Foundations for Modelling Uncertainties?

Henry E. Kyburg
kyburg@cs.rochester.edu
Department of Computer Science
University of Rochester
Rochester, New York, 14627

## 1 What Are Foundations?

Surely we want solid foundations. What kind of castle can we build on sand? What is the point of devoting effort to balconies and minarets, if the foundation may be so weak as to allow the structure to collapse of its own weight? We want our foundations set on bedrock, designed to last for generations. Who would want an architect who cannot certify the soundness of the foundations of his buildings?

The architectural analogy is not entirely persuasive, though. It is easy to understand what the foundations of a building are: they are the steel pilings, the stone, the concrete, on which the rest of the edifice physically stands. It stands still. Science is not like that. It is living, dynamic, constantly changing. The foundations of a part of science are themselves open to change.

So what are the foundations of a branch of science? There are the first principles, that everyone can and does (at least for the time being) agree on. The axioms of the discipline. Often these foundations are provided by a different branch of science. Thus engineering takes for granted — uses as a foundation — statics, kinematics, dynamics... In that sense the foundations of uncertainty clearly include logic and mathematics: we don't need to invent the real number system for ourselves.

But there may be something more we want, something more domain specific, on which we can build our science. Some key idea, like the idea of 'force' in physics, or of a 'neighborhood' in topology. We have yet to agree on any such idea or set of ideas: we argue with each other and interrogate our intuitions. And disagree at length. We would liked to find relations, common elements, among the diverse ideas with which we work. But there is no unified theory, no general framework to which we can turn to see the relation between one idea and another.

At the same time, though, we do manage to put together useful systems in which uncertainty figures. In fact people with intense and widely divergent views regarding the nature of uncertainty have been successful in making practical use of their ideas regarding uncertainty, its measurement, procedures for updating uncertainty, its role in decision making, etc. How is this possible? How can you do something right if you start off on the wrong foot? Clearly, since each of n people thinks the other n - 1 are wrong, not more than one can be proceeding correctly. And yet they are getting at least part way to their goals.

## 2 Do We Need Foundations At All?

In this situation it is easy enough to see why some people are willing to say that we should not bother with foundations at all. A similar situation can be found in logic and mathematics: there are those who *do* mathematics, and then there are those who do a different thing — the *foundations* of mathematics. "It's all very interesting," one might say, "to speculate about foundations, but it's not *serious* work."

In fact this attitude can be raised to the level of a matter of principle. Of course we should be explicit about what our models contain, but it need not be the case that there is any privileged class of models. Different models are useful for different things. To seek a unification of or a foundation for modelling uncertainties is, it might be argued, completely misguided. It can only inhibit free and untrammeled and creative research. "Let a thousand flowers bloom!"

There is a movement in the philosophy of science that leads in the same direction. It can be traced to Paul Feyerabend and Thomas Kuhn [Feyerabend, 1970, Kuhn, 1952], and the gist of the view is that science is driven by practical and also by social goals. The community of scientists accepts a certain 'disciplinary matrix' that embodies rules of evidence, criteria of sound argument, and the like, but these standards change over time. When it comes to science itself, there is no truth of the matter. Scientific argument, for example about the conservation of parity or the measurement of belief, is a formal dance, conforming to temporary rules of argument.

This is a perennial view. A number of years ago, the idea was that there were always 'presuppositions,' so that if you made your presuppositions explicit, you had exercised full foundational responsibility. If my presuppositions differ from yours, why, we have no disagreement after all: one would not expect to reach the same substantive conclusions on the basis of differing presuppositions.

So too, more recently, with the idea of a 'model.' If I argue that the evidence supports the hypothesis that people are expected utility maximizers, and you argue that the evidence doesn't either support that hypothesis, what we must do is to sit down and carefully characterize the models we are respectively employing. Lo and behold: we find that they are not the same model, so of course we are not really disagreeing. What is true on your model is false on mine, and vice versa.

What is nice about all these anti-foundational ploys is that they promote friendship and collegiality; they can be used very effectively to resolve disagreements. After all, if we are using different models of uncertainty, why should we expect to come to the same conclusions?

## 3 Testability

There are a number of difficulties with the soft friendly world of relativism. For one thing, it conflicts with the persistent nagging feeling that, by gum, there *is* a truth of the matter out there. We know there is, because we have knocked into tables, fallen out of trees, gotten rained on. Reality isn't all bad. We have also basked in the sun, made love, seen magnificent sunsets. The point is that those things, good and bad, are not social constructs: they are the world going about its business.

The view that we must answer to the world as well as to our colleagues is an old and honorable one. Rudolf Carnap, among others, proposed *testability* as a criterion for meaningfulness in the theory of language. [Carnap, 1934] Construed narrowly, the idea of testability is implausible: there are many statements in science that we cannot definitively test. Any universal generalization provides an illustration: we cannot complete an inventory of the universe. Construed broadly, testability might be thought of as the demand that any meaningful statement should be such that you can find evidence supporting it. This begins to get pretty vague.

Karl Popper proposed *falsifiability* as a criterion of meaningfulness. It didn't fare much better. But in both cases, it is assumed that there is a truth of the matter in these controversial issues, and one that is worth pursuing.

## 4 Proliferation and Communication

The idea that we want to encourage uninhibited creativity in formulating theories and devising structures to represent the world and the activities of intelligent entities in the world is not at all peculiar to the anti-foundational view. Popper and Carnap, for example, both stressed the importance of speculation and invention. The question is partly that of whether or not there are objective standards for evaluating speculations and inventions.

The question is also partly one of communication: it is not only the acceptability or unacceptability of some statement or other, but a question of the very meaning of that statement. If you and I have different paradigms (or models, or assumptions) then when you assert that our robot should be designed to maximize expected utility, and I assert that it should not merely maximize expected utility, we can agree that our disagreement is only apparent, and derives from the difference in our assumptions. When you assert that our robot will avoid risk, how am I to understand that? Without a common framework, how do we even *express* our differences?

In choosing among alternative frameworks, the external world provides a powerful input, according to traditional views. What has been less noticed is that without a shared framework that is tied to the world, we cannot communicate. We can smile and nod, but the empirical value of what you say is lost on me.

Unbounded proliferation, the proliferation of private languages applicable, to all intents and purposes to private worlds, is clearly counterproductive. We can't get very far that way, because we can't communicate and cooperate. That is one reason why we need to consider foundations.

## 5 Considering Foundations

What is it to 'consider foundations?' One might think it a matter of consulting one's intuitions, and trying to formalize them. Perhaps that is one approach, and if pursued with an awareness of both the history of the subject, and with an alertness to applications in the real world, it can be very interesting.

There are other ways of thinking about foundations that involve looking carefully at ready-made systems. For example, authors often argue that other systems are "special cases" of their own. Shafer [Shafer, 1976] shows that Bayesian propositional probabilities are a special case of his belief functions. It goes the other way, too. Dempster [Dempster, 1965] mentions the possibility that his upper and lower probabilities could be construed as the envelope of a set of classical probability functions, and cites Savage as having suggested that these could be considered convex sets.

The same general view has been endorsed by Isaac Levi [Levi, 1984] under the rubric of "indeterminate probabilities."

David Heckerman [Heckerman, 1985] has shown that MYCIN's certainty factors should (if they are to be fed into decision procedures) be construed in a manner consistent with propositional probabilities.

Updating uncertainties in the light of evidence is a quite different matter. It can be shown that in many cases updating belief functions results in tighter constraints on uncertainty measures than does updating according to Bayes' theorem applied to propositional probabilities [Kyburg, 1989].

One way in which one might want to resolve some of these questions is to adopt the principle of maximizing expected utility as a constraint: that is, it should *not* be the case that the expected utility of an admissible act, under any point-valued probability function that is a *possibility*, is less in every state of nature than the expected utility of some other act.

These results concern the relation of convex sets of probability measures to certainty factors or belief functions. This suggests that one might, more deeply, consider the question of whether convex sets of probability functions are the most useful general representation of uncertainty. Grounds for a negative answer can be found (implicitly) in Pearl's work [Pearl, 1989]. Consider a chance event such as the tossing of a coin. We could be uncertain as to the bias of the coin, but at the same time quite certain as to the independence of the tosses. Since the convex combination of two binomial distributions of heads is not a binomial distribution, we do not want to identify rational belief with

Another foundational issue is whether Bayes' theorem should be applied to the updating of probabilities. This is considered in [Dubois, 1988] as well as [Kyburg, 1990].

## 6 What Are We Trying to Do?

Different people have different ideas of the goal of modelling uncertainty. Some people are mainly concerned with the representation of human cognition. Others want machines that will handle an uncertain environment with efficiency and grace. These may be two quite different projects (though, presumably, related in some degree). They may call for somewhat different foundational considerations.

For example, if we are interested in how people manage to deal with uncertainty, the idea of a connectionist network may be a very useful one: perhaps that is the basic mechanism we need to embody.

On the other hand, if we are interested in machines that will deal with statistically known uncertain environments, then it may well be that a different foundation is called for: one which depends on a database of statistical knowledge, rather than on being trained by the environment.

The contrast is that between programming a machine to do logic, and constructing a network (which has been done) that will learn logic. I understand that a network has been constructed that learns logic at about the "B" level. It is easy to construct a program that will do logic much better. Which you are interested in depends on your goals. I'm more interested in a pocket calculator that does sums correctly than in one that gets a "B" in third grade arithmetic, but does it the way children do.

## 7 What Are We Talking About?

One reason for concern with foundations in modelling uncertainty is to enable us to know what we are talking about. If we don't have some foundational framework, some commonality, communication is undermined. If you use 'probability' in one sense and I use it in another, it is clear that we're in for problems. One response has been to proliferate terminology. Thus 'epistemic probability' 'belief function' 'basic probability assignment' 'certainty factor' 'support' ...

But the proliferation of terminology is not necessarily the proliferation of ideas, and does not always enhance communication. It is true that it may help to eliminate or reduce *mis*communication: I won't misinterpret you if you are talking your own language, but avoiding misunderstanding is not the same as successfully communicating.

Is communication so important? Might we proceed along an evolutionary model, in which each of us pursues his own dream, designs his own systems, and eventually, is evaluated by the success of his work? Should we let the thousand flowers bloom, and then let all but one wither? This doesn't seem like science. (Furthermore, if all but one is to die, we'd better make sure that it is self fertile.)

## 8 Little but the Truth

If, as I think most of us really believe, there is a truth of the matter about the nature (or natures!) of uncertainty, about its representation, and about its modification by evidence or the course of experience (not necessarily the same thing), then foundational issues are clearly important. The closer we can come to an adequate foundation, the more likely we are to be telling the truth when we talk about uncertainty.

This is not at all to say that there is a final foundation, any more than there is a final foundation to physics or to mathematics. It is to say that our work concerning uncertainty will be less likely to contain error. What we pass on will be more likely to be permanent.

## 9 More of the Truth

We're not merely interested in avoiding falsehoods, however; we are interesting in telling the truth. A better understanding of foundations cannot help but provide insights from which more new results can be obtained. This is particularly the case when the foundational work has shown that two apparently different concepts are simple transformations of one another. Every theorem about the one becomes, with the appropriate transformations, a theorem about the other.

For example, once we have noted that every belief function is equivalent to a convex set of classical probabilities, then theorems concerning convex sets of classical probabilities can be applied to belief functions.

## 10 Usefulness

We have already noted that, despite the strong divergence of opinion concerning foundations, individuals in a wide variety of camps have managed to produce useful systems, ranging from concrete applications in expert systems to purely abstract shells for representing uncertain inference. This strongly suggests that there is an interesting foundational core underlying those applications.

It does not *entail* that there is a unified foundation; maybe there are $k$ foundations for $k$ kinds of uncertainty. That's just one of the foundational issues that deserves careful and generous investigation.

If that is the case, however, if these successful applications reflect a unified common core of facts about uncertainty, then having a better understanding of the foundations will provide us with tools for doing things even better.

This observation goes both ways. The more useful applications there are of a certain sort, the stronger the evidence that there is something right in what is being done. This fact should be exploited in the study of foundations. Its exploitation can be neither mechanical nor trivial, however. It is all too easy to take one framework, and to show how a second can be paraphrased in terms of the first.

This is especially the case for the subjectivist Bayesian view. If there is a system that yields decisions on the basis of evidential inputs, and if as I suggested decision procedures should not *conflict* with probabilities, then it is necessarily the case that there are subjective prior probabilities that yield the same decisions.

Does this mean that subjective Bayesianism is the Universal Solvent? Hardly, because there are other desiderata than conformity to the probability calculus that are or may be part of our search for foundations.

## 11 Practise and Theory

This suggests that practise and theory should go hand in hand. We can engineer applications involving uncertainty that work. We can do so starting from a variety of theoretical and philosophical viewpoints. In view of that variety, we can hardly claim support, on this basis, for any particular foundational view.

On the other hand, the space of possible foundational theories is very large. There is a lot more to be thought about than is involved in any particular engineering development. We must also consult our intuitions — and those of others — in order to have reason to believe that our engineering successes are not accidents.

This doesn't mean that foundational work is a matter of sitting in an armchair thinking deep thoughts. Philosophers are traditionally good at that. But unless the deep thoughts are tested, we can't tell what they have to offer us. Unless the intuitions obtained in the armchair can be put to the test of practise, they can have little persuasiveness.

Practise and theory must go together; to get very far, they must go together systematically, not just accidentally.

## 12 A Garden

A concern with foundations may be in some small degree conservative — at least in comparison with the radical view of Feyerabend. But there is no one that I know of, in any area, who takes the view that creativity and the proliferation of ideas are to be discouraged. Foundations, even where they exist, are not immutable and unchangeable.

Nor does there have to be uniform agreement on the foundations of uncertainty modelling. The more uniformity we can find, the more agreement we can achieve, the better. A common framework allows us to communicate better, it allows us to transfer results from one part of the theory to another, and we may hope that eventually it will allow us to understand better why the systems that work do work.

Finding a common framework within which many of the current ideas concerning the representation and updating of uncertainty can be related need not inhibit the proliferation of ideas. If we're lucky, it will enhance their effect by enhancing their communication.

Let a thousand flowers bloom; but let them bloom together and cooperatively in a formal garden.

## Acknowledgments

Research underlying this work has been supported in part by U. S. Army Communications Command Grant no. DAAB10-87-K-022, and NSF research grant no. IRI-9002659.